\documentclass[runningheads]{llncs}
\usepackage[T1]{fontenc}
\usepackage{graphicx}
\usepackage{amsmath, amssymb}
\usepackage{microtype}
\usepackage{multirow}
\usepackage{tcolorbox}
\usepackage{enumitem}
\usepackage{xcolor}

\begin{document}
\title{Multimodal Graph Retrieval-Augmented Sequential Recommendation via Collaborative Filtering Paths}
\titlerunning{MGRASRec}
% If the paper title is too long for the running head, you can set
% an abbreviated paper title here
%
\author{Jason Marcell Setiadi \and Xin Cao \and Lina Yao}
\authorrunning{J. M. Setiadi et al.}
% First names are abbreviated in the running head.
% If there are more than two authors, 'et al.' is used.
%
\institute{University of New South Wales, Sydney, Australia\\
\email{jason.setiadi@unsw.edu.au, xin.cao@unsw.edu.au, lina.yao@unsw.edu.au}}
\maketitle              % typeset the header of the contribution
\begin{abstract}
Multimodal Large Language Models (MLLMs) have demonstrated strong potential for sequential recommendation through their ability to reason over complex multimodal data. However, existing approaches either rely solely on the target user's own interaction history, neglecting collaborative signals from neighboring users, or incur substantial computational overhead through repeated MLLM inference over long interaction histories. To address these challenges, we propose MGRASRec, a multimodal graph retrieval-augmented framework for sequential recommendation. MGRASRec injects collaborative filtering signals conditioned on the candidate item directly into the MLLM prompt by retrieving structured paths from a user-item interaction graph, extended via multimodal similarity to increase coverage beyond exact co-interaction overlap. This retrieval also surfaces the history items most relevant to the candidate at no additional cost, removing the need for recurrent summarization and keeping inference to a single forward pass per candidate. All components are unified into an augmented prompt for parameter-efficient fine-tuning of an MLLM. Extensive evaluations across three publicly available datasets validate the effectiveness of MGRASRec, achieving the best performance on all metrics with particularly strong gains in ranking quality.

\keywords{Multimodal Sequential Recommendation \and Multimodal Large Language Model \and Retrieval-Augmented Generation}
\end{abstract}

% 1. What is the problem? Why is it interesting and important?
% 2. What are the core challenges? Why is it hard?
% 3. Why hasn’t it been solved before? What are the limitations of previous solutions? (no mention or critiquing papers in intro, leave it to related works)
% 4. What are the key components of my approach and results? Also include any specific limitations.

\section{Introduction}
Sequential recommendation (SR) in real-world applications inherently involves rich multimodal item information (images, videos, audio, and textual metadata), and leveraging these signals enables a deeper understanding of user preferences beyond purely ID-based representations \cite{liu2024mllmsurvey}. Recent advances in Multimodal Large Language Models (MLLMs) have significantly improved multimodal comprehension by unifying diverse inputs into a shared semantic space, enabling stronger reasoning over complex multimodal data across many domains. Motivated by these advances, recent works have adapted LLMs and MLLMs for sequential recommendation \cite{lin2025llmsurvey,wu2024survey}, either as semantic encoders to generate item representations \cite{wang2024llmrg,zhang2025notellm} or leveraging their reasoning capabilities directly for recommendation generation \cite{geng2022p5,liu2024recgpt}, showing particular promise for capturing user preferences and high-level semantic relationships across interaction sequences.

Despite this promise, two critical challenges remain unresolved. First, existing MLLM-based approaches operate in isolation from the broader behavioral patterns of similar users. Collaborative filtering signals are a well-established driver of recommendation quality \cite{he2020lightgcn}, yet integrating them into prompt-based MLLM frameworks remains an open problem \cite{wang2025mmsrarec}. The difficulty is compounded by sparsity, as identifying such signals through co-interaction requires exact item overlap between users, leaving a substantial fraction of user-item pairs without any collaborative signal.
Second, sequential recommendation naturally involves interaction histories that accumulate over time, making it infeasible to place every item in the MLLM prompt. A common workaround is to truncate history based on recency or frequency, but this may discard items most relevant to the candidate. While recurrent summarization approaches have been proposed to preserve the full history \cite{wang2025mllmsrec,ye2025mllmmsr}, they introduce substantial computational overhead through repeated MLLM inference, making them impractical for real-world deployment.

We propose MGRASRec, which addresses both challenges with a single design decision. MGRASRec injects candidate-conditioned collaborative signals directly into the MLLM prompt by retrieving structured paths over the user-item interaction graph that connect the target user to the target item via bridge users, extended with a multimodal similarity layer to widen coverage beyond exact co-interaction overlap. As a byproduct, the same retrieved paths surface the history items most structurally relevant to the candidate, enabling path-guided history selection at no additional inference cost. 
We perform parameter-efficient fine-tuning of an MLLM as a recommender, using an augmented prompt that integrates the retrieved collaborative paths and selected history items. We evaluate MGRASRec on three public datasets from various domains, where it outperforms strong baselines on all metrics. Our main contributions are summarized as follows:
\begin{itemize}
    \item We inject candidate-conditioned collaborative filtering signals into MLLM-based sequential recommendation via structured graph path retrieval, which ablations identify as the primary driver of performance gains across three datasets.
    \item We introduce path-guided history selection, which surfaces candidate-relevant history items as a zero-cost byproduct of path retrieval, keeping recommendation to a single MLLM forward pass per candidate.
\end{itemize}

\section{Related Work}
\label{sec:related_works}

The strong semantic understanding of LLMs and MLLMs has motivated growing interest in recommendation. One line of work uses LLMs and MLLMs as feature encoders. LLMRG \cite{wang2024llmrg} and KAR \cite{xi2024kar} augment conventional recommenders by extracting structured reasoning and factual knowledge from LLMs, while NoteLLM-2 \cite{zhang2025notellm} and MLLMRec \cite{dang2025mllmrec} employ MLLMs to transform multimodal item information into unified semantic representations. Molar \cite{luo2024molar} further incorporates collaborative filtering signals with multimodal content through post-alignment contrastive learning. However, these approaches decouple semantic understanding from the recommendation decision, meaning the model's reasoning capabilities are never directly applied to the final prediction.

A complementary line directly formulates recommendation as a natural language generation task. P5 \cite{geng2022p5} reformulates diverse recommendation tasks as prompt-based text generation, and TALLRec \cite{bao2023tallrec} applies parameter-efficient fine-tuning to adapt LLaMA for recommendation. Extending this paradigm to the multimodal setting, MLLM-MSR \cite{ye2025mllmmsr} recurrently summarizes user interaction histories into natural language preference descriptions through multiple rounds of MLLM inference, which makes recommendation-time cost grow linearly with history length. MMSRARec \cite{wang2025mmsrarec} injects collaborative signals by retrieving embedding-similar users and surfacing their subsequent interactions as keyword context, but this signal is not conditioned on the candidate item, as bridge users are selected based on general history similarity rather than whether they actually interacted with the target item, providing unfocused collaborative evidence. Our work addresses these limitations by introducing structured graph-based collaborative retrieval that encodes behavioral evidence conditioned on the candidate item, a direction that remains largely unexplored in prompt-based MLLM frameworks.

\begin{figure*}[!t]
  \centering
  \includegraphics[width=\textwidth]{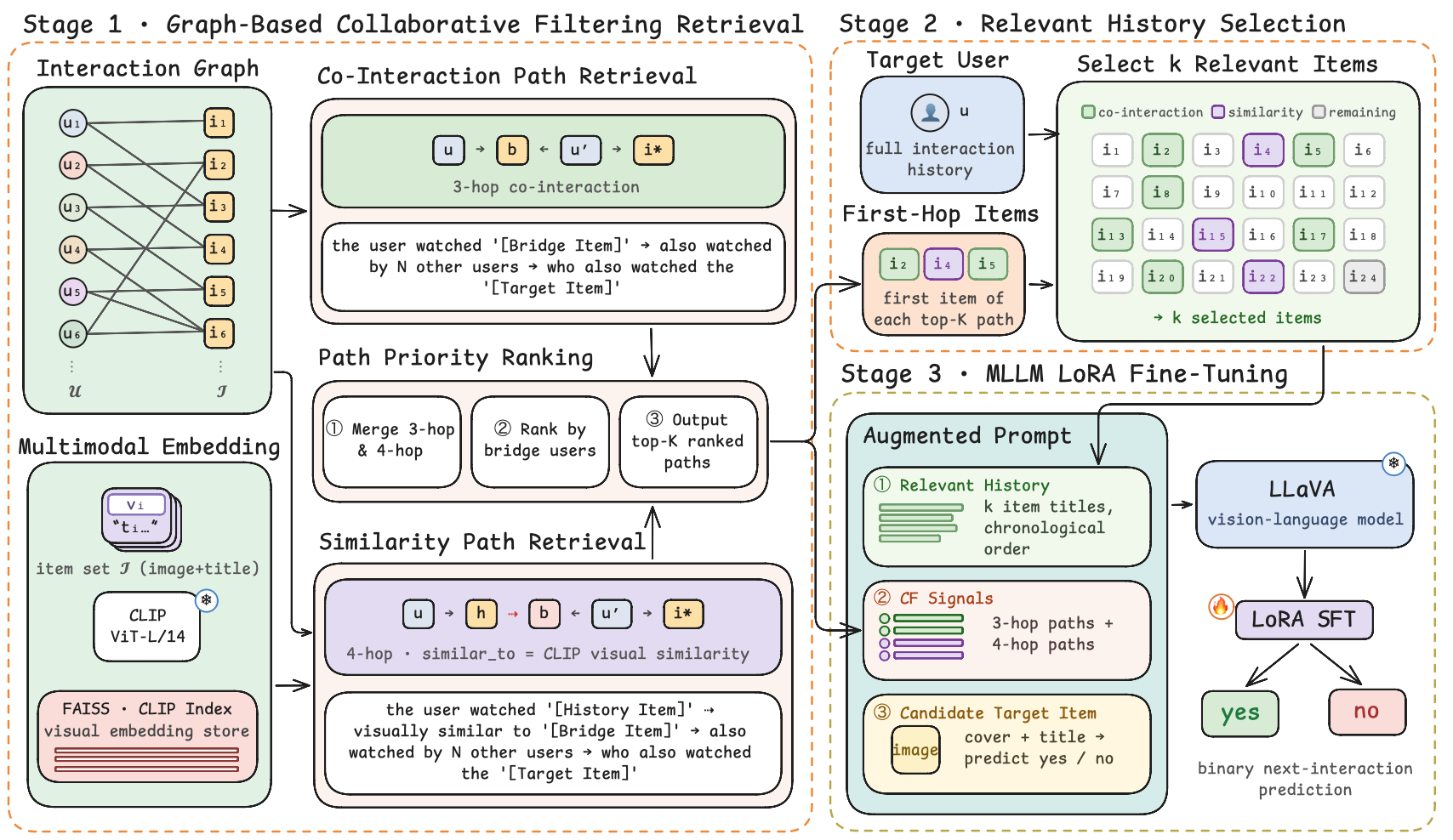}
  \caption{The framework of MGRASRec}
  \label{fig:framework}
\end{figure*}

\section{Method}

In this section, we introduce a three-stage framework illustrated in Fig.~\ref{fig:framework}, comprising graph-based collaborative filtering retrieval, relevant history selection, and MLLM LoRA fine-tuning. We formalize the problem setting in Section~\ref{subsec:prob}, before detailing each stage.

\subsection{Problem Formulation}
\label{subsec:prob}
We formally define the multimodal sequential recommendation problem as follows. Given a user $u$, we define their historical behavior sequence as $S_u = [I_1^u, \ldots, I_n^u]$, where $I_i^u$ represents the $i$-th item with which the user has interacted, and $n$ denotes the length of the sequence. Each item $i$ in the item catalog $\mathcal{I}$ is associated with a textual title $t_i$ and an image $v_i$ (e.g., product image or video cover). Given the historical behavior sequence $S_u$ and a candidate item $I_c$, the objective of multimodal sequential recommendation is to predict the probability that $I_c$ is the user's next interaction, i.e., $P(I^u_{n+1} = I_c \mid S_u)$.

\subsection{Graph-based Collaborative Filtering Retrieval}
\label{subsec:cf}
To inject focused collaborative signals conditioned on the candidate item directly into the recommendation prompt, we represent the user-item interaction history as a bipartite graph $\mathcal{G} = (\mathcal{U} \cup \mathcal{I},\ \mathcal{E})$, where an undirected edge $(u, i) \in \mathcal{E}$ indicates that user $u$ has interacted with item $i$. When verbalizing paths in the prompt, each edge is expressed directionally as (user, watched, item) or (item, watched\_by, user), with the verb instantiated per domain (`watched' for Microlens, `bought' for Amazon). It is built offline over the interaction history, excluding each split's held-out positive pairs to avoid leakage. All path retrieval described below is likewise performed offline, decoupled from serving. We then retrieve paths over $\mathcal{G}$ connecting the target user to the target item via bridge users, as detailed below.

\begin{table}[h]
\caption{Path retrieval coverage (\%) for positive (ground-truth) and negative (sampled) candidates, reported as Train/Val/Test}
\begin{center}
\setlength{\tabcolsep}{4pt}
\begin{tabular}{|l|l|c|c|}
\hline
\textbf{Dataset} & \textbf{Path Type} & \textbf{Positive} & \textbf{Negative} \\
\hline
\multirow{3}{*}{Microlens}
& Co-interaction only & 66.4/71.0/69.2 & 17.0/21.4/19.4 \\
& Similarity only     & 99.1/99.2/100.0 & 88.8/91.0/88.0 \\
& Combined            & \textbf{99.3/99.4/100.0} & \textbf{89.1/91.4/88.5} \\
\hline
\multirow{3}{*}{Amazon-Games}
& Co-interaction only & 73.8/77.2/78.8 & 27.6/29.8/31.6 \\
& Similarity only     & 98.5/98.8/99.2 & 86.2/89.0/88.8 \\
& Combined            & \textbf{98.6/98.8/99.6} & \textbf{86.6/89.0/89.2} \\
\hline
\multirow{3}{*}{Amazon-Baby}
& Co-interaction only & 71.0/71.6/69.2 & 28.7/34.0/30.7 \\
& Similarity only     & 98.3/99.6/98.4 & 93.8/95.6/94.6 \\
& Combined            & \textbf{98.5/99.6/98.4} & \textbf{94.3/96.0/94.9} \\
\hline
\end{tabular}
\label{tab:coverage}
\end{center}
\end{table}

\subsubsection{Co-Interaction Path Retrieval}
For each (target user $u$, target item $i^*$) pair, we perform a breadth-first search (BFS) over $\mathcal{G}$ to retrieve 3-hop co-interaction paths of the form:
\begin{equation}
    u \xrightarrow{\text{watched}} b \xleftarrow{\text{watched}} u'
    \xrightarrow{\text{watched}} i^*
\end{equation}
where $b \in \mathcal{I}$ is a bridge item co-interacted by both $u$ and a bridge user $u'$, and $u'$ has also interacted with the target item $i^*$.
This 3-hop structure is the shortest path connecting the target user to the target item through a bridge user, capturing the most direct collaborative evidence possible, compared to longer paths that would pass through additional intermediate users. Concretely, we retrieve $S_u$ and the users who have interacted with $i^*$, and identify shared items between the target user and each bridge user as valid bridge items, each yielding one path, verbalized in the prompt as shown in Fig.~\ref{fig:mllm_prompt} (with the number of bridge users capped for efficiency at scale). However, this approach requires exact co-interaction overlap, which leaves a significant fraction of user–item pairs without any collaborative signal, especially for negative candidates (Table~\ref{tab:coverage}). To extend coverage, we additionally accept bridge items that are visually or semantically similar to items already in the target user's history.

\subsubsection{Similarity Path Retrieval}
To identify such bridge items, we first compute multimodal embeddings for all items using CLIP (ViT-L/14)~\cite{radford2021clip}, where for each item $i$, the image embedding and title embedding are averaged and L2-normalized to yield a fused item representation:
\begin{equation}
    \mathbf{e}_i = \mathrm{normalize}\!\left(
        \frac{\mathrm{CLIP}_{\mathrm{img}}(v_i) + \mathrm{CLIP}_{\mathrm{txt}}(t_i)}{2}
    \right)
\end{equation}
All item embeddings are stored offline in a FAISS \texttt{IndexFlatIP} index, enabling efficient cosine similarity search via inner product on the normalized vectors.
For each pair $(u, i^*)$, we batch-query the FAISS index for visually similar items per history item in $S_u$ whose cosine similarity exceeds the 99th percentile of each dataset's similarity distribution, excluding items already in $S_u$ to avoid duplicating co-interaction paths. For each bridge user $u'$ that has interacted with the target item $i^*$, we check whether $u'$ has also interacted with any of these similar items, and if so emit a 4-hop similarity path of the form:
\begin{equation}
    u \xrightarrow{\text{watched}} h \xrightarrow{\text{similar\_to}} b
    \xleftarrow{\text{watched}} u' \xrightarrow{\text{watched}} i^*
\end{equation}
where $h \in S_u$ is a history item and $b$ is the similar bridge item, verbalized analogously (Fig.~\ref{fig:mllm_prompt}). The final hop remains strict, requiring bridge user $u'$ to have directly interacted with the target item $i^*$, preserving the quality of the collaborative signal while widening the set of valid bridge items through visual and semantic similarity. Identifying these paths requires an ANN search over the precomputed FAISS index followed by a BFS traversal over $\mathcal{G}$ to verify the final co-interaction hop. Table~\ref{tab:coverage} confirms that incorporating similarity paths increases path coverage beyond co-interaction alone, covering almost all user-item pairs across all three datasets for both positive and negative candidates.

\subsubsection{Path Priority Ranking}
\label{sec:merging}
Since the number of retrieved paths grows unboundedly with interaction density, including all of them would result in excessive prompt length and introduce noise from weakly-supported paths. We therefore select only top-$K$ paths while including a count summary of total available co-interaction and similarity paths in the prompt, preserving signal breadth without additional token cost, as illustrated in Fig.~\ref{fig:mllm_prompt}. We sweep $K\in\{3,5,7,10\}$ and observe no significant performance difference, and fix $K=5$ as a balance between prompt length and coverage. Paths are ranked by the number of unique bridge users passing through each bridge item, reflecting consensus behavioral evidence. Co-interaction paths are prioritized with at least 60\% of top-K slots reserved as they encode more direct collaborative evidence (Section~\ref{subsec:ablation}), while remaining slots are filled by the next highest-ranked paths regardless of type.

\subsection{Relevant History Selection}
\label{subsec:history}

A key challenge in MLLM-based sequential recommendation is handling long user interaction histories. Providing the full history in the prompt is infeasible, while recurrent MLLM inference for summarization incurs substantial computational overhead. A natural alternative is to select a subset of the most relevant history items, but naively truncating by recency or frequency provides no guarantee that selected items relate to the target item, as the most relevant interactions may lie far from the tail of the sequence for users with long histories. We address this with a path-guided selection that surfaces the k history items most structurally relevant to the target item at zero additional inference cost. Since path retrieval identifies which history items are connected to the target item via the first hop of each retrieved path, this selection emerges directly as a byproduct of retrieval already performed. Concretely, we extract the first-hop history item from each of the top-K ranked paths, following the same co-interaction priority logic as Section~\ref{sec:merging}. For sparse or no-path cases, remaining slots are padded with the most frequent items from the interaction history not already selected, ensuring a non-empty history in all cases. The final selection is restored to chronological order, preserving the sequential structure the model was pretrained to process.
 
\subsection{MLLM LoRA Fine-Tuning}
\label{subsec:finetuning}

\begin{figure}[t]
\centering
\begin{tcolorbox}[
    colback=white,
    colframe=black,
    width=\linewidth,
    boxrule=1pt,
    arc=0pt,
    top=2pt,
    bottom=4pt,
    left=4pt,
    right=4pt,
    boxsep=0pt
]
\scriptsize
\noindent
\begin{minipage}[c]{0.75\linewidth}
\textbf{\#\#\# User:}\\
\texttt{<image>} The user's interaction history:
\begin{enumerate}[noitemsep, topsep=2pt]
    \item \textcolor{red}{April new: excellent students and Miss daughter \# set it photosynthesis creator \# spring storm hit \# cuckoo's engagement}
    \item \textcolor{red}{you can question his strength but not his appearance level \# Sasuke}
    \item \textcolor{red}{\# Quadratic Nova \# Comic recommendations \# Comic commentary}
    \item[] \textcolor{red}{\ldots}
\end{enumerate}
Collaborative filtering signals (\textcolor{red}{6 co-interaction, 6 similarity-based}):
\begin{itemize}[noitemsep, topsep=2pt]
    \item \textcolor{red}{the user watched `you can question his strength but not' $\rightarrow$ also watched by 4 other users $\rightarrow$ who also watched `some promises can only be lies in'}
    \item \textcolor{red}{the user watched `I usually ignore people bothering you because' $\rightarrow$ visually similar to `I will always love you \# Naruto' $\rightarrow$ also watched by 2 other users $\rightarrow$ who also watched `some promises can only be lies in'}
    \item[] \textcolor{red}{\ldots}
\end{itemize}
Please predict whether this user would interact with the video at the next opportunity. The video's title is \textcolor{red}{`some promises can only be lies in the end \# Whirlpool Naruto \# Sun to Daisada \# summer strike'}, and the given image is this video's cover. Please only respond `yes' or `no'.
\end{minipage}%
\hfill
\begin{minipage}[c]{0.22\linewidth}
    \centering
    \includegraphics[width=\linewidth]{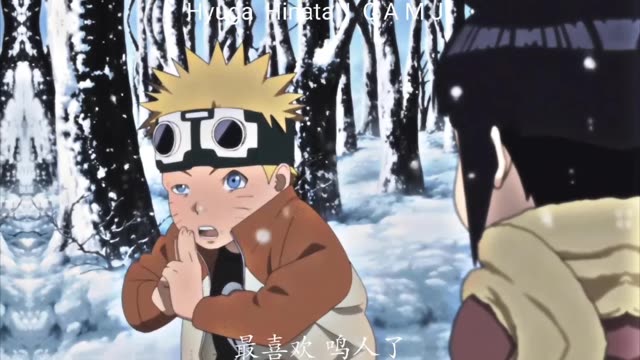}
\end{minipage}
\tcblower
\scriptsize
\textbf{\#\#\# Assistant:} yes
\end{tcolorbox}
\caption{An example prompt template of MLLM-based sequential recommendation from the Microlens dataset.}
\label{fig:mllm_prompt}
\end{figure}

The retrieved collaborative signals and selected history items are assembled into an augmented prompt, illustrated in Fig.~\ref{fig:mllm_prompt}. If no paths exist for a given pair, the CF signals section is omitted and the prompt contains only the history and target item. 
We apply supervised fine-tuning (SFT) of LLaVA \cite{liu2023llava}, where the training objective follows the next-token prediction paradigm, and the same augmented prompt format is used at both training and inference time, ensuring no distributional shift between the two stages. We adopt LoRA~\cite{hu2022lora} under the PEFT framework to reduce training cost while preserving the model's in-context reasoning capability. At inference time, the probability $P(I^u_{n+1} = I_c \mid S_u)$ is estimated from the first generated token:
\begin{equation}
    p = \frac{p(\text{`yes'})}{p(\text{`yes'}) + p(\text{`no'})}
\end{equation}
At recommendation time, MGRASRec requires only this single forward pass per candidate. We quantify the resulting efficiency advantage over MLLM-MSR \cite{ye2025mllmmsr} in Section~\ref{subsec:efficiency}.

\section{Experiments}

\subsection{Experimental Setup}
\subsubsection{Dataset Description}
We evaluate on three open-source, real-world datasets spanning diverse recommendation domains: (1) Microlens \cite{ni2025microlens}, a micro-video recommendation dataset containing user-item interactions, video introductions, and cover images; (2) Amazon-Games \cite{hou2024amazon}, reflecting user preferences in the digital goods sector; and (3) Amazon-Baby \cite{hou2024amazon}, from the e-commerce domain representing purchasing behavior in the baby product category. All datasets comprise user-item interactions, product descriptions, and images.
We applied k-core filtering with thresholds of (7, 5) for Microlens and (6, 5) for Amazon, iteratively removing users and items below the minimum interaction count. Due to MLLM context window limitations, we limit each user's history to 5 items for Microlens and 10 for Amazon, sorted chronologically. Users were randomly partitioned into training, validation, and test sets at an 8:1:1 ratio. For negative sampling, we applied a 1:1 positive-to-negative ratio for training and validation, and a 1:20 ratio for testing. Negatives are sampled uniformly at random from items the target user has not interacted with.

\subsubsection{Baseline Methods}
To evaluate the effectiveness of MGRASRec, we compare against representative baselines across four categories. For \textbf{basic SR models}, GRU4Rec \cite{hidasi2016gru4rec} uses Gated Recurrent Units to model sequential dependencies between items, and SASRec \cite{kang2018sasrec} employs self-attention to capture long-term user preferences, both relying solely on item IDs and collaborative signals. For \textbf{multimodal recommendation models}, MMGCN \cite{wei2019mmgcn} integrates multimodal features into a graph-based framework via message passing, while FREEDOM \cite{zhou2023freedom} leverages modality-specific graph structures and edge pruning strategies to improve robustness and mitigate modality noise. For \textbf{LLM-based SR models}, TALLRec \cite{bao2023tallrec} fine-tunes an LLM for sequence recommendation via supervised fine-tuning using exclusively textual item information. Finally, for \textbf{MLLM-based SR models}, we evaluate LLaVA both zero-shot and LoRA fine-tuned with frequency-based history truncation and no collaborative signals, equivalent to the w/o All Components variant in Section~\ref{subsec:ablation}. MLLM-MSR \cite{ye2025mllmmsr} converts item images and text into natural language descriptions via an MLLM and infers user preferences through multiple rounds of reasoning with supervised fine-tuning.\footnote{The concurrent preprint MMSRARec \cite{wang2025mmsrarec} is excluded from quantitative comparison as it has no released code, and its dataset preprocessing and evaluation protocol differ from ours, making reported numbers incomparable. We discuss it qualitatively in Section~\ref{sec:related_works}.}

\subsubsection{Evaluation Metrics}
We evaluate under a ranking-stage protocol, where the model re-scores a shortlist consisting of the ground-truth positive and sampled negative candidates rather than searching the full item catalog, consistent with the evaluation convention adopted by the baselines we compare against. To assess performance under this protocol, we employ AUC, HR@5, and MRR@5 as evaluation metrics for baseline methods and our proposed method.
% To assess the performance of baseline methods and our proposed method for multimodal sequential recommendations, we employ AUC, HR@5, and MRR@5 as evaluation metrics. 

\subsubsection{Implementation Details}
% on the KATANA High Performance Computing Cluster \cite{katana} 
All experiments were conducted on a single NVIDIA H200 GPU. We use LLaVA-v1.6-Mistral-7B as the backbone MLLM, fine-tuned with LoRA (rank 16, alpha 32, dropout 0.1) using the AdamW optimizer with a learning rate of 2e-5, a per-device batch size of 1, gradient accumulation of 4 steps, and 4 training epochs. All baselines are evaluated on the same preprocessed datasets, data splits, and evaluation protocol. For MLLM-MSR we use the authors' official implementation under the same SFT settings as ours.

\begin{table}[!t]
\caption{The performance of different methods.}
\begin{center}
\resizebox{\textwidth}{!}{%
\begin{tabular}{|l|c|c|c|c|c|c|c|c|c|}
\hline
& \multicolumn{3}{c|}{\textbf{Microlens}} & \multicolumn{3}{c|}{\textbf{Amazon-Games}} & \multicolumn{3}{c|}{\textbf{Amazon-Baby}} \\
\cline{2-10}
& \textbf{AUC} & \textbf{HR@5} & \textbf{MRR@5} & \textbf{AUC} & \textbf{HR@5} & \textbf{MRR@5} & \textbf{AUC} & \textbf{HR@5} & \textbf{MRR@5} \\
\hline

GRU4Rec & 73.02 & 51.92 & 26.19 & 76.54 & 61.82 & 32.77 & 70.20 & 49.57 & 24.23 \\
SASRec  & 74.49 & 57.48 & 29.40 & 74.15 & 46.31 & 22.15 & 77.02 & 61.45 & 40.76 \\
MMGCN   & 73.82 & 53.07 & 26.97 & 78.00 & 58.13 & 33.40 & 71.40 & 48.49 & 24.55 \\
FREEDOM  & 79.62 & 66.40 & 46.40 & 80.12 & 65.20 & 43.89 & 77.72 & 64.40 & 45.73 \\
TALLRec  & 85.34 & 73.87 & 52.22 & 90.33 & 86.40 & 65.27 & 82.91 & 70.80 & 48.86 \\
LLaVA    & 51.07 & 25.60 & 13.89 & 55.07 & 30.40 & 15.26 & 52.27 & 30.00 & 14.99 \\
LLaVA w/ SFT & 87.98 & 81.07 & 58.42 & 90.81 & 86.13 & 66.62 & 83.51 & 74.40 & 53.03 \\
MLLM-MSR & 82.16 & 74.00 & 48.30 & 83.61 & 73.73 & 53.72 & 78.10 & 62.00 & 46.48 \\
\hline
MGRASRec & \textbf{89.55} & \textbf{83.07} & \textbf{62.65} & \textbf{91.96} & \textbf{89.47} & \textbf{69.94} & \textbf{85.54} & \textbf{77.40} & \textbf{56.59} \\

\hline
\end{tabular}
}
\label{tab:performance}
\end{center}
\end{table}

\subsection{Overall Performance}

The performance of all compared methods is presented in Table~\ref{tab:performance}, averaged over 3 independent runs. MGRASRec achieves the best performance on all metrics across all three datasets. Multimodal models (FREEDOM, MMGCN) and ID-only sequential models (GRU4Rec, SASRec) each capture only part of the relevant signal and generally underperform fine-tuned MLLM-based approaches, underscoring that integrating both sequential and multimodal processing is essential. Among LLM/MLLM-based models, supervised fine-tuning is critical, with LLaVA w/ SFT substantially outperforming zero-shot LLaVA. Notably, MLLM-MSR underperforms both LLaVA w/ SFT and MGRASRec despite sharing the same backbone and SFT setup, which we attribute to item-level information loss from compressing history into a high-level preference summary rather than preserving item-specific signals via chronological prompting. Building on this fine-tuned MLLM, MGRASRec further injects candidate-conditioned collaborative signals directly into the prompt, yielding the strongest performance overall with particularly pronounced gains on the ranking metrics HR@5 and MRR@5, indicating that collaborative signals capture preference information complementary to sequential and multimodal modeling alone.

\subsection{Ablation Study}
\label{subsec:ablation}

\begin{figure*}[!t]
  \centering
  \includegraphics[width=\textwidth]{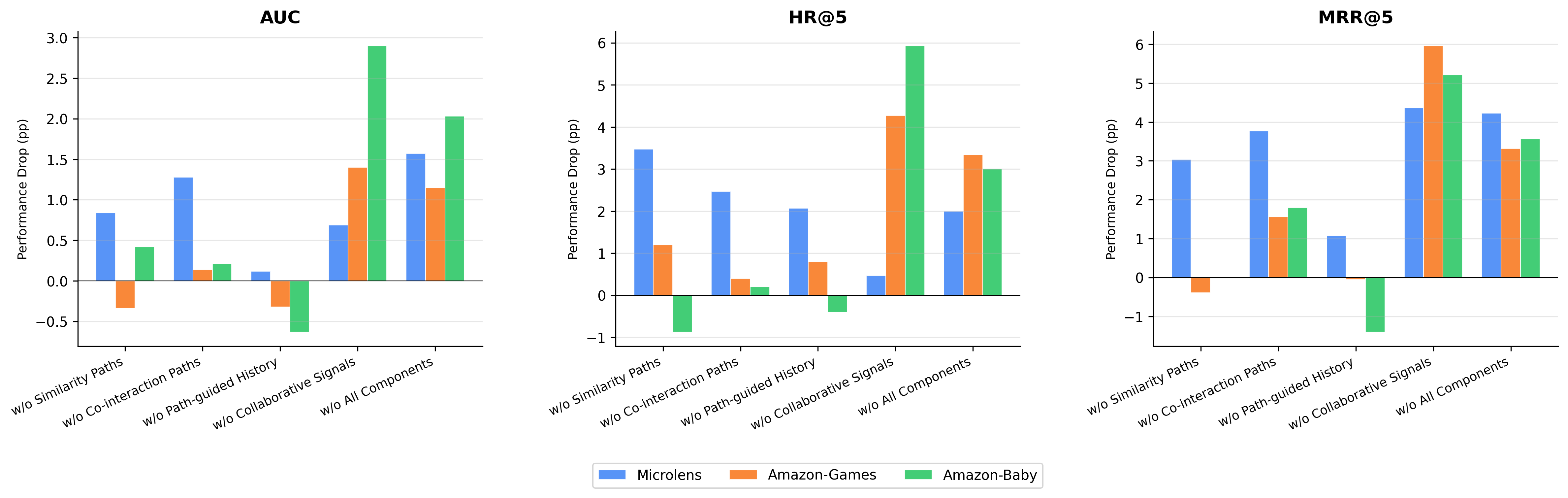}
  \caption{Performance drop (in percentage points) when each component is removed from MGRASRec, across three datasets and three metrics. Larger bars indicate greater contribution of the removed component. Negative values indicate cases where removal improves performance, reflecting dataset-specific variation in component effectiveness.}
  \label{fig:ablation}
\end{figure*}

To evaluate the contribution of each component, we ablate MGRASRec against five variants. \textbf{w/o Similarity Paths} removes the CLIP-based retrieval layer, retaining only co-interaction paths. \textbf{w/o Co-interaction Paths} removes 3-hop co-interaction retrieval, retaining only similarity paths. \textbf{w/o Path-guided History} replaces path-guided selection with frequency-based truncation. \textbf{w/o Collaborative Signals} removes both similarity and co-interaction paths from the prompt. \textbf{w/o All Components} reduces the model to a vanilla SFT baseline with frequency-based history truncation and no collaborative signals.

Results are shown in Fig.~\ref{fig:ablation}, reporting performance drop in percentage points relative to the full model. Across all three datasets, collaborative filtering signals are consistently the most impactful component, though the relative contribution of co-interaction versus similarity paths varies by domain. Co-interaction paths are the more reliable driver on the Amazon datasets, while similarity paths matter more on Microlens, where visual semantics correlate more strongly with user interest. Path-guided history selection shows a dataset-dependent effect on accuracy, improving all three metrics on Microlens but inconsistent or slightly negative elsewhere. Its primary value lies in efficiency, since it surfaces relevant history items at no extra cost given the retrieved collaborative paths.

\subsection{Efficiency Analysis}
\label{subsec:efficiency}

\begin{figure*}[!t]
  \centering
  \includegraphics[width=0.5\textwidth]{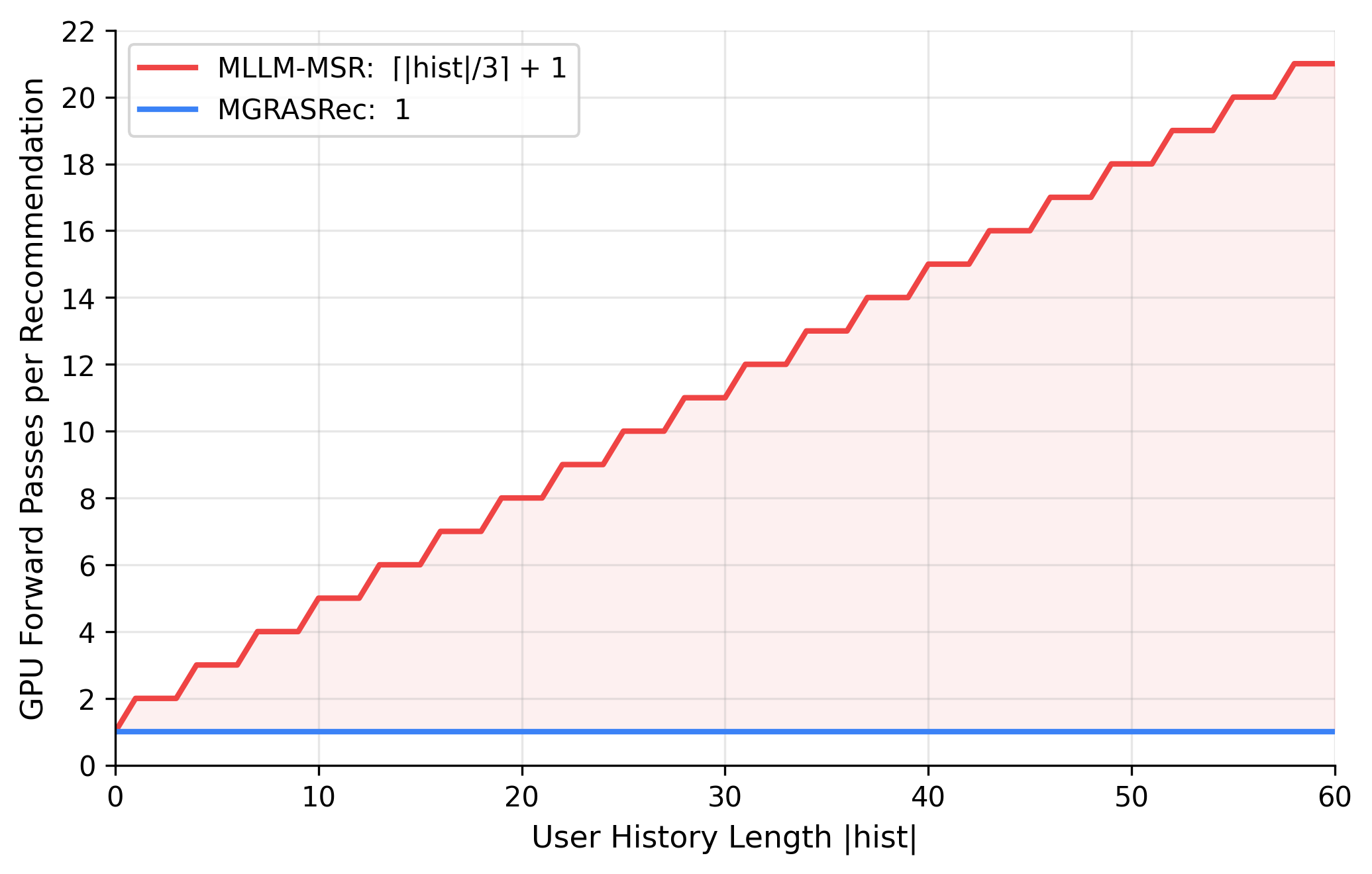}
  \caption{GPU forward passes required at recommendation time as user history scales. MLLM-MSR incurs $\lceil |hist| / 3 \rceil$ additional LLM forward passes for recurrent preference summarization. MGRASRec requires exactly one regardless of history length.}
  \label{fig:efficiency}
\end{figure*}

We analyze inference cost at recommendation time for MGRASRec and MLLM-MSR. As shown in Fig.~\ref{fig:efficiency}, inference cost grows linearly with user history length for MLLM-MSR but remains constant for MGRASRec. This is achieved by shifting retrieval offline, where path construction requires only a BFS traversal over the interaction graph and lookups into a precomputed item-similarity table, both CPU-bound operations that are inexpensive relative to an MLLM forward pass. Note that this constant cost is with respect to history length only, yet the total number of MLLM forward passes still equals the number of candidate items scored, as MGRASRec evaluates each candidate independently via a single binary yes/no query.

\section{Conclusion}
We propose MGRASRec, a multimodal graph retrieval-augmented framework for sequential recommendation that addresses key limitations of existing MLLM-based approaches. By injecting structured collaborative signals from neighboring users directly into the MLLM prompt via graph path retrieval, extending coverage through multimodal similarity-based bridge items, and selecting candidate-relevant history directly from the retrieved paths, MGRASRec leverages shared behavioral patterns beyond individual user history at a constant inference cost. Our results demonstrate that augmenting MLLMs with structured graph-based collaborative paths is the primary driver of MGRASRec's gains, indicating that this is a promising direction for multimodal sequential recommendation. 
Currently, our formulation frames recommendation as binary prediction over a constrained candidate set, requiring one forward pass per candidate item. Future work could inject graph-based collaborative signals into open-set generation, where the MLLM directly generates the next item from the full item vocabulary under constrained decoding for item grounding.

\bibliographystyle{splncs04}
\bibliography{mybibliography}
%
% \begin{thebibliography}{8}
% \bibitem{ref_article1}
% Author, F.: Article title. Journal \textbf{2}(5), 99--110 (2016)

% \bibitem{ref_lncs1}
% Author, F., Author, S.: Title of a proceedings paper. In: Editor,
% F., Editor, S. (eds.) CONFERENCE 2016, LNCS, vol. 9999, pp. 1--13.
% Springer, Heidelberg (2016). \doi{10.10007/1234567890}

% \bibitem{ref_book1}
% Author, F., Author, S., Author, T.: Book title. 2nd edn. Publisher,
% Location (1999)

% \bibitem{ref_proc1}
% Author, A.-B.: Contribution title. In: 9th International Proceedings
% on Proceedings, pp. 1--2. Publisher, Location (2010)

% \bibitem{ref_url1}
% LNCS Homepage, \url{http://www.springer.com/lncs}, last accessed 2023/10/25
% \end{thebibliography}
\end{document}